\documentclass[10pt, conference]{IEEEtran}
\IEEEoverridecommandlockouts
\usepackage{cite}
\usepackage{amsmath,amssymb,amsfonts}
\usepackage{algorithmic}
\usepackage{graphicx}
\usepackage{color}
\usepackage{textcomp}
\usepackage{tabularx}
\usepackage{multirow}
\usepackage{comment}
\usepackage{listings}
\usepackage{url}
\usepackage[T1]{fontenc}
\usepackage{array}
\usepackage{booktabs}
\usepackage{float}

\newcommand\shimari[1]{}
\newcommand\todo[1]{}

\def\BibTeX{{\rm B\kern-.05em{\sc i\kern-.025em b}\kern-.08em
    T\kern-.1667em\lower.7ex\hbox{E}\kern-.125emX}}
\begin{document}

\title{Round Outcome Prediction in VALORANT\\ Using Tactical Features from Video Analysis\thanks{Short Paper}}

\author{\IEEEauthorblockN{
Nirai Hayakawa\IEEEauthorrefmark{1}, 
Kazumasa Shimari\IEEEauthorrefmark{2}, 
Kazuma Yamasaki\IEEEauthorrefmark{2}, \\
Hirotatsu Hoshikawa\IEEEauthorrefmark{2}, 
Rikuto Tsuchida\IEEEauthorrefmark{2}, 
Kenichi Matsumoto\IEEEauthorrefmark{2}, 
}
\IEEEauthorblockA{\IEEEauthorrefmark{1} Miyazaki Nishi High School, Miyazaki, Japan, 
Email: exam.niraihayakawa@gmail.com}
\IEEEauthorblockA{\IEEEauthorrefmark{2} Graduate School of Science and Technology, Nara Institute of Science and Technology, Nara, Japan, \\
Email: \{k.shimari, yamasaki.kazuma.yj9, hoshikawa.hirotatsu.hh4, tsuchida.rikuto.tq5, matsumoto.kenichi\}@naist.ac.jp}
}

\maketitle
\begin{abstract}
Recently, research on predicting match outcomes in esports has been actively conducted, but much of it is based on match log data and statistical information.
This research targets the FPS game VALORANT, which requires complex strategies, and aims to build a round outcome prediction model by analyzing minimap information in match footage.
Specifically, based on the video recognition model TimeSformer, we attempt to improve prediction accuracy by incorporating detailed tactical features extracted from minimap information, such as character position information and other in-game events. 
This paper reports preliminary results showing that a model trained on a dataset augmented with such tactical event labels achieved approximately 81\% prediction accuracy, especially from the middle phases of a round onward, significantly outperforming a model trained on a dataset with the minimap information itself.
This suggests that leveraging tactical features from match footage is highly effective for predicting round outcomes in VALORANT. 

\end{abstract}


\section{Introduction}
\label{sec:introduction}
Esports have gone beyond mere entertainment to become established as a global competitive field, with its market size continually expanding. 
Particularly, First-Person Shooters (FPS) have captivated large audiences as a popular genre, demanding advanced strategy, teamwork, and precise individual player skill. 
However, due to their complexity, it is not easy for audiences, especially those less experienced, to understand in real-time how subtle in-match situational changes and player decisions connect to win or lose.
As a result, it can be difficult for those unfamiliar with the game to follow the action~\cite{Hamari2017IR}.
This is one of the challenges for the further popularization and expansion of the esports fan base.


To address this issue, real-time match outcome prediction in esports has been adopted to enrich the audience experience. 
Many existing studies in Multiplayer Online Battle Arena (MOBA) and other genres of games utilize log data or statistics~\cite{hucapp22}\cite{hodge2021win}\cite{CBIC2023-161}.
This information is useful for understanding the game state and predicting the outcome of the game, 
However, it is difficult to fully capture the nuances of on-screen visual information, especially the rich tactical details consolidated on the map.
In FPS games like VALORANT, visual information such as the positions of teammates and opponents on the minimap strongly affect the outcome of a match. 
This makes it important to analyze visual information related to the minimap.

Another approach involves the direct analysis of match footage.
Chulajata et al.\cite{chulajata2025Realtime} targeted the 2D fighting game Super Street Fighter II Turbo, performing real-time outcome prediction using only the time-series changes in both players' health bars as input for LSTM and Transformer Encoder models. 
This research highlights the importance of using real-time video data for analysis, demonstrating its potential to enhance audience engagement in esports.
Although this method extracts information from match footage, FPS games require consideration of more complex tactical features.

In this paper, we propose a method for round outcome prediction in VALORANT by analyzing match footage, with a particular focus on visual tactical features from the minimap using deep learning. 
We adopt TimeSformer~\cite{bertasius2021space} as the base model due to its strength in understanding spatio-temporal contexts. 
This approach uses visual tactical features from the minimap and predicts round outcomes in VALORANT. These tactical features, such as inferred enemy positions from auditory cues and the strategic use of skills, have the potential to enable more nuanced predictions.
Our goal is to achieve more accurate outcome predictions by analyzing match footage with tactical features.


The remainder of this paper is organized as follows. 
Section~\ref{sec:background} reviews related work and positions our study. 
Section~\ref{sec:proposed_method} details the proposed method. 
Section~\ref{sec:result} describes the experimental setup and evaluation results. 
Section~\ref{sec:discussion} discusses the findings, and Section~\ref{sec:conclusion} concludes the paper.

\section{Background}
\label{sec:background}
\subsection{Match Outcome Prediction in esports}
Match outcome prediction in esports games has become an active research area. 
A major approach involves utilizing statistical information obtainable from match log data or APIs.
In the Multiplayer Online Battle Arena (MOBA) genre, there is some research targeting Dota 2~\cite{hodge2021win}\cite{yang2017real}. 
They analyzed log data such as kills/deaths and gold differences during matches in real time and estimated win rates with high accuracy using logistic regression and ensemble learning. 
In LoL, Junior et al.\cite{CBIC2023-161} performed win/loss predictions using API data from various time slices and LightGBM.
In the FPS genre CS:GO, Makarov et al.\cite{makarov2017predicting} and Xenopoulos et al.\cite{xenopoulos2020valuing} predicted match and round outcomes based on log data using logistic regression and XGBoost, respectively. 
These studies demonstrate the effectiveness of predictive models using structured data. 
However, it is not always possible to use all the necessary information from logs and APIs, and there may be important information displayed on the screen that cannot be collected.

Another approach involves evaluating player performance from past match data and predicting future match outcomes based on it.
Bahrololloomi et al.\cite{hucapp22} proposed a method to calculate a performance score for each player from past LoL data using machine learning and predict match outcomes by considering this score and the player's role.
This is interesting in that it models player abilities, but it does not directly reflect real-time match situations.

Research on directly analyzing real-time match footage is also progressing, but its direct application to win/loss prediction is still limited.
Chulajata et al.\cite{chulajata2025Realtime} performed real-time win/loss prediction for the 2D fighting game Super Street Fighter II Turbo using the time-series changes in both players' health bars as input for LSTM and Transformer Encoder models. 
While this extracts information from match footage, it uses only health bars' information and differs from the more complex spatial and tactical information from FPS minimaps targeted in this study.

\subsection{Match Outcome Prediction using Tactical Features}

To improve the accuracy of win/loss prediction, it is important to consider not only simple statistics like kill counts but also tactical features such as player positioning, movement, and team formations. In MOBAs, Rioult et al. \cite{rioult2014mining} and Deja et al.\cite{deja2015topological} demonstrated the importance of topological and geographical features of teams.
In CS:GO, Xenopoulos et al. \cite{xenopoulos2022esta} constructed a spatio-temporal dataset containing detailed player behavior logs and developed a high-precision prediction model using it.
In VALORANT, the use of tactical features is also being attempted, such as DeRover et al.\cite{derover2023winning} predicting 1v1 combat outcomes based on player coordinates. 
These studies emphasize the importance of tactical features but are mainly based on extraction from log data. 
This research aims to advance existing approaches by extracting these important tactical features from real-time match footage.

\subsection{Video Recognition using Transformers}
The recent success of Transformers in deep learning has also spread to the field of video recognition. The Transformer, proposed by Vaswani et al. \cite{NIPS2017_3f5ee243}, excels at capturing long-range dependencies in sequential data through its self-attention mechanism.
One model that applies this to video is TimeSformer~\cite{bertasius2021space}, a Transformer-based architecture designed for video understanding. 
TimeSformer, through innovations such as factorized self-attention mechanisms that separate spatial and temporal attention, can learn complex spatio-temporal features in videos more effectively and with higher computational efficiency compared to traditional 3D CNNs. It is expected to be particularly effective for tasks like ours, which involve long temporal sequences and where inter-frame relationships are important.
As an application to game video analysis, Joo et al. \cite{joo2025improving} used ViT for object recognition on game screens, suggesting the effectiveness of Transformer-based models. In this research, we apply TimeSformer to the task of round outcome prediction in VALORANT and verify its effectiveness.


\section{Proposed Method}
\label{sec:proposed_method}
We propose a method that predicts the round outcome in VALORANT based on match footage. 
We hypothesize that explicitly modeling the tactical features will lead to improved prediction accuracy. 
To test this hypothesis, we design two variants of our model with different input feature sets and train them on datasets constructed from official VALORANT tournament broadcasts. 
We then analyze their predictive performance as well as the importance of individual features.

\subsection{Model Architecture}
As the foundation of our predictive model, we adopt TimeSformer \cite{bertasius2021space}, a video recognition model based on the Transformer architecture. TimeSformer applies the Transformer framework to sequential video data, enabling it to efficiently capture both temporal dependencies and spatial features. Compared to conventional 3D convolutional neural networks, TimeSformer is not only more memory-efficient on GPUs but also capable of handling significantly longer video sequences, which makes it suited for our task.

\subsection{Dataset Construction}
\subsubsection{Dataset Collection}
We collect broadcast video from official VALORANT tournaments held in 2024.
The total number of collected videos is 1,376, and the number of rounds is 29,506.
Since a standard VALORANT round lasts up to 100 seconds, excluding overtime, only rounds that concluded within this time limit are included in the dataset. A total of 21,229 rounds satisfy this condition.
The distribution of these round durations is as follows: 254 rounds last between 0 and 25 seconds, 2,682 rounds between 26 and 50 seconds, 7,966 rounds between 51 and 75 seconds, and 10,327 rounds between 76 and 100 seconds.

Additionally, the match footage collected in this study is cropped from the tournament broadcast. 
Therefore, while the minimap in normal match footage only shows the player's team's position and skills, the tournament match footage displays the positions and skill information for both teams on the minimap.

\subsubsection{Dataset}

To train and evaluate our models, we constructed the following two datasets:

    \subsection*{Dataset A: Minimap information}
     From each video, we extract the minimap in the game and segment the video into individual rounds. 
     Each round is labeled with its outcome (win or loss) and the map on which it was played. 
     This dataset offers raw minimap information, such as the positions of characters.
     Round segmentation is automated using EasyOCR~\cite{easyocr}, which is one of the most popular lightweight OCR tools, and labeling is conducted based on collected round outcomes.
    
    \subsection*{Dataset B: Minimap Information Augmented with Tactical Events}
    Building on Dataset A, we augment the data with additional labels using image processing techniques. Specifically, we use OpenCV-based template matching to identify and label tactical events. 
    These events typically signify when and where teams gain or lose tactical information, often derived from visual or auditory cues within the game environment. 
    In our target game, VALORANT, such tactical events include the detection of enemy presence through auditory cues like footsteps; the fixed audible range of these footsteps helps determine if an opponent moving at a certain speed is heard.
    These events also cover the strategic implications of skill usage, which involves monitoring skill icons and their effects to label moments of intelligence gained or denied. 
    Therefore, the resulting dataset captures both direct and inferred tactical events crucial for understanding strategic decision-making.
    


\section{Evaluation}
\label{sec:result}
We evaluate our developed AI model by comparing the prediction accuracy of two models: Model A, trained on Dataset A (minimap labeled only with win/loss outcomes), and Model B, trained on Dataset B, which includes additional tactical event labels.
\subsection{Evaluation Method}
We train separate TimeSformer-based models using Dataset A and Dataset B, resulting in Model A and Model B, respectively.
The models are implemented in PyTorch. 
We adopt a TimeSformer architecture using its default settings, which include 12 layers, 12 attention heads, and divided space-time attention. 
The embedding dimension is uniformly set to 768 across all layers. Input videos are downsampled to 8 frames per second (fps).
To prevent overfitting, we apply a dropout rate of 0.1. Training is conducted with early stopping, terminating when the validation accuracy does not improve for 30 consecutive epochs.
In Dataset B, each event label consists of a timestamp, team side, agent name, and area name. 
These features are aggregated using average pooling over every 8-frame segment. 
The pooled features are then passed through an embedding layer followed by a linear projection to reduce their dimensionality. 
The resulting event representations are early fused with visual features extracted from the corresponding video frames. 
These fused representations are input into the TimeSformer model in 8-frame chunks.
For optimization, we use the AdamW optimizer with a learning rate of 1e-4 and weight decay of 1e-4.  
A linear warm-up is applied over the first 5,000 steps, followed by a cosine annealing learning rate schedule.
Model performance is evaluated on a separate validation set, using the same preprocessing and inference pipeline as during training. 
Predictions are generated at one-second intervals throughout each round. 

As the evaluation metric, we use the per-second accuracy of round outcome predictions. 
We specifically analyze how prediction accuracy varies over the course of each round, aiming to identify when and what types of information most strongly contribute to correct predictions.
The validation set is curated by segmenting the round from start to finish and assigning ground-truth outcome labels automatically. 
For this study, we randomly sample 100 rounds from the dataset, selecting only those with complete and uninterrupted match sequences.

\subsection{Result}

\begin{table}[tb]
  \centering
  \caption{Average Prediction Accuracy(\%)}
  \label{tab:result_table}
  \begin{tabular}{lrrrrr}
    \toprule
    \textbf{Model}    & \textbf{Overall} & \textbf{0–24s} & \textbf{25–49s} & \textbf{50–74s} & \textbf{75–99s} \\ 
    \midrule
    Model A        & 72.28  & 62.48 & 72.88 & 73.88 & 79.88             \\ 
    Model B        & 80.55 & 56.32  & 85.80 & 90.64 & 89.44            \\
    \bottomrule
  \end{tabular}
\end{table}

\begin{figure}[tb]
\centering
\includegraphics[width=\linewidth]{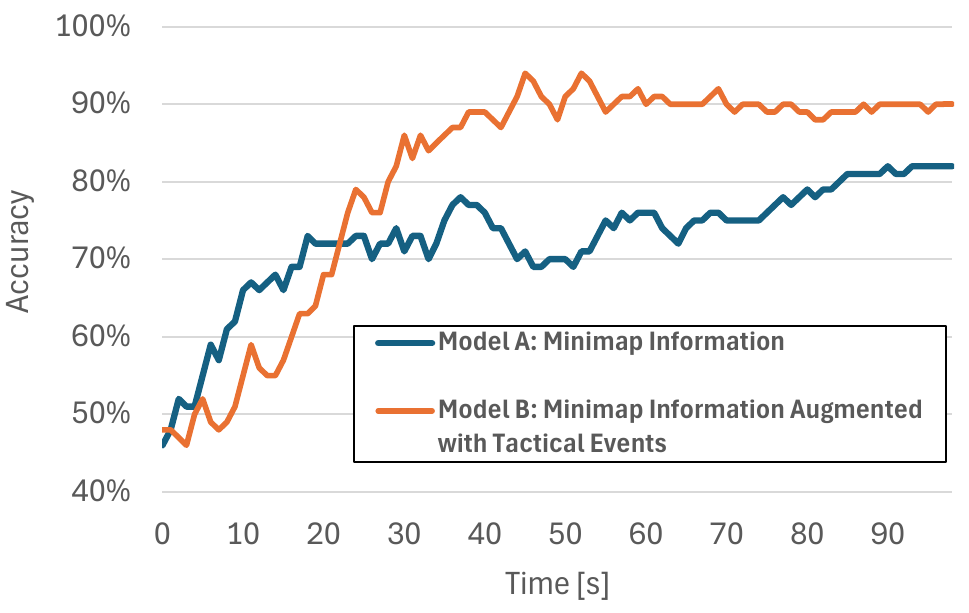}
\caption{Prediction Accuracy}
\label{fig:prediction_accuracy}
\end{figure}

Table~\ref{tab:result_table} shows the average prediction accuracy of each model in specific time intervals within a round. The overall refers to the average obtained by summing the model's prediction accuracies for each second.
The overall prediction accuracy across the entire round reached approximately 80.55\% for Model B, which has additional tactical events labeled. 
This result is approximately 8 percentage points higher than Model A, which only uses minimap information.

Figure~\ref{fig:prediction_accuracy} also shows how prediction accuracy changes over time from the start of each round.
The results show that model B consistently outperforms model A from approximately 24 seconds after the start of the round.
Notably, when focusing on the data from 30 seconds into each round, the accuracy increases to over 80\%.
In VALORANT, key abilities that significantly affect the course of a round are often activated from the middle of the round. 
The results suggest that the additional temporal and spatial information utilized by Model B plays an important role in interpreting game states following the deployment of abilities. In particular, information that enables inference of ability usage contributes to the improved prediction accuracy observed in the latter part of each round.
The superior performance of Model A in the early phase of the round may be due to the absence of additional events given to Model B during that phase.


\section{Discussion} 
\label{sec:discussion}
This section discusses the main findings of the study, along with its limitations and potential directions for future work.

The superior prediction accuracy demonstrated by the model trained on Dataset B, particularly during the mid-to-late phases of each round, suggests that detailed tactical event labels, including spatial information and inferred skill usage, are critical features for accurately predicting round outcome in VALORANT. 
This highlights the importance of tactical features extracted from match footage that cannot be captured by simple minimap information.



Furthermore, while our method has achieved a certain level of accuracy, there is room for further improvement.
In the current implementation, predictions are made based on minimap information augmented with tactical events. 
Therefore, our implementation does not consider player-specific characteristics such as individual past performance data, as well as time-series information like movement speed and direction derived from changes in coordinates, which can represent player intentions and tactical movements. 
This information could further improve the accuracy of round outcome predictions.

\section{Conclusion}
\label{sec:conclusion}
In this study, we propose a method for round outcome prediction in the esports title VALORANT. 
Leveraging TimeSformer as the backbone architecture, we trained and evaluated two models: one on Dataset A, which includes only minimap and round-level outcome labels, and another on Dataset B, which incorporates additional spatial and tactical event labels extracted from the minimap.
Experimental results demonstrate that Model B, which leverages tactical features, achieved substantially higher prediction accuracy, reaching approximately 80\% from 30 seconds into each round. 
These findings suggest that detailed in-game visual and tactical features play a critical role in predicting round outcomes, particularly in the mid-to-late phases of tactically complex matches.

We aim to advance this work by enriching input feature representations, such as movement vectors and player-specific historical data, and refining the model architecture. 
Furthermore, we aim to extend this system to provide coaching for non-professional players by analyzing and explaining effective strategies and decision-making processes of professional players, fostering a deeper in-game strategic understanding.

\section*{Acknowledgement}

This research has been supported by JSPS KAKENHI No. JP23K16862 and JST Science and Technology Challenge Program for Next Generation.

\bibliographystyle{IEEEtran.bst}
\bibliography{IEEEabrv,reference}
\end{document}